\documentclass[conference,a4paper]{IEEEtran}
\IEEEoverridecommandlockouts

\usepackage{cite}
\usepackage{amsmath,amssymb,amsfonts}
\usepackage{algorithmic}
\usepackage{graphicx}
\usepackage{textcomp}
\usepackage{booktabs}
\usepackage{color,soul}
\usepackage{tabularx,hhline}
\usepackage{multirow}
\usepackage{hyperref}
\usepackage{tikz}
\usepackage{textcomp}
\usepackage{xcolor}
\usepackage{cancel}
\usepackage{enumitem}

\usepackage[strings]{underscore}

\begin{document}

\title{Physics-Informed Neural Networks for Accelerating Power System State Estimation}
\author{
\IEEEauthorblockN{Solon Falas}
\IEEEauthorblockA{\textit{ECE Dept., KIOS CoE}\\
\textit{University of Cyprus}\\
Nicosia, Cyprus \\
falas.solon@ucy.ac.cy}
\and
\IEEEauthorblockN{Markos Asprou}
\IEEEauthorblockA{\textit{ECE Dept., KIOS CoE} \\
\textit{University of Cyprus}\\
Nicosia, Cyprus \\
asprou.markos@ucy.ac.cy}
\and
\IEEEauthorblockN{Charalambos Konstantinou}
\IEEEauthorblockA{\textit{CEMSE Division} \\
\textit{KAUST}\\
Thuwal, Saudi Arabia \\
charalambos.konstantinou@kaust.edu.sa}
\and
\IEEEauthorblockN{Maria K. Michael}
\IEEEauthorblockA{\textit{ECE Dept., KIOS CoE} \\
\textit{University of Cyprus}\\
Nicosia, Cyprus \\
mmichael@ucy.ac.cy}
}
\IEEEaftertitletext{\vspace{-2\baselineskip}}

\IEEEoverridecommandlockouts

\maketitle

\IEEEpubidadjcol

\begin{abstract}
State estimation is the cornerstone of the power system control center, since it provides the operating condition of the system in consecutive time intervals. This work investigates the application of physics-informed neural networks (PINNs) for accelerating power systems state estimation in monitoring the operation of power systems. Traditional state estimation techniques often rely on iterative algorithms that can be computationally intensive, particularly for large-scale power systems. In this paper, a novel approach that leverages the inherent physical knowledge of power systems through the integration of PINNs is proposed. By incorporating physical laws as prior knowledge, the proposed method significantly reduces the computational complexity associated with state estimation while maintaining high accuracy. The proposed method achieves up to 11\% increase in accuracy, 75\% reduction in standard deviation of results, and 30\% faster convergence, as demonstrated by comprehensive experiments on the IEEE 14-bus system.

\end{abstract}

\vspace{-0.5\baselineskip}
\section{Introduction}\label{sec:Introduction}
Power system state estimation is crucial for reliable and secure grid operation. However, traditional techniques relying on SCADA measurements suffer from sparse and error-prone data, leading to delayed and less accurate estimates.
Conventional state estimation techniques, rely on complex iterative methods, which are prone to large delays in case of large scale power systems. Furthermore, if the measurement set of the state estimation includes both conventional measurements (i.e., power flow/injection) and measurements from Phasor Measurement Units (PMUs), large delays might affect the monitoring responsiveness of the state estimator for capturing short-duration transients.
Machine learning approaches offer accelerated state estimation by processing measurements promptly after neural network training. This paper introduces physics-informed neural networks (PINNs) to meet the need for faster state estimation. While responsiveness may not be critical for conventional measurements with low reporting rates, high PMU observability in future power systems necessitates accelerated state estimation to leverage real-time PMU reporting.

Many of the existing machine learning techniques if used for power system state estimation demonstrate drawbacks in capturing the complex dynamics and constraints of power systems\cite{mukherjee2022power}. Relying solely on statistical patterns in historical data can lead to inaccurate estimates, particularly in scenarios with limited or noisy data. Moreover, training traditional learning models requires substantial amounts of data, which is costly to collect in power system applications. These limitations underscore the need for advanced approaches that combine machine learning with domain-specific physics knowledge to improve the accuracy and reliability of state estimation.

PINNs have emerged as a promising approach for solving complex problems in various scientific and engineering domains \cite{falas2020water,raissi2019physics}, including power system state estimation \cite{huang2022applications}. PINNs offer several advantages over traditional machine learning techniques. Firstly, PINNs integrate domain-specific physical laws and constraints into the Neural Network (NN) architecture, enabling the incorporation of prior knowledge about the system behavior. This ensures that learned models are consistent with underlying physics, leading to more accurate and interpretable results. Secondly, PINNs can effectively handle data scarcity by leveraging physics-based regularization terms, reducing the reliance on large datasets. Additionally, PINNs enable the efficient handling of multi-dimensional inputs and outputs, making them suitable for complex power system modeling and control tasks. The flexibility and interpretability of PINNs make them valuable for various use-cases, including power system parameter estimation, fault detection and diagnosis of power system operation.

In this paper, a novel NN training method for power system state estimation by leveraging the physics-informed approach is presented. The proposed approach integrates the physical laws and constraints of power systems as prior knowledge into the NN training process. The performance of the proposed architecture is tested under various training scenarios, comparing it against a benchmark plain NN. The results demonstrate that the proposed PINN achieves higher accuracy, improved algorithmic stability, and requires less training effort compared to the benchmark model. These findings highlight the potential of PINNs as a powerful accelerator for power system state estimation, particularly in the context of the PMU era, where real-time and accurate estimation is crucial for ensuring reliable power system operation.

The remainder of this paper is organized as follows: Section \ref{sec:Related Work} provides a review of related work in the field of state estimation and the application of NNs. Section \ref{sec:Methodology} presents the methodology employed in this study, which includes a background on PINNs and the formulation of the training process. Section \ref{sec:Experimental Setup & Results} presents the experimental results, while Section \ref{sec:Conclusion} concludes the paper.
\section{Related Work}\label{sec:Related Work}
Traditional power system state estimation techniques, such as the Gauss-Newton and Weighted Least Squares methods, have long relied on iterative approaches and measurements obtained from legacy sensors transmitted through SCADA systems. However, these methods can become computationally demanding when applied to large-scale power systems. Additionally, the sparse and infrequent nature of measurements from legacy sensors can result in delays and decreased accuracy, especially during dynamic system conditions. These limitations highlight the need for more advanced techniques that can overcome these challenges and provide more efficient and accurate state estimation in power systems.

In the past, efforts were focused on utilizing machine learning techniques, including artificial NNs and support vector machines, to overcome the limitations of traditional state estimation methods. By using historical data and statistical patterns, these approaches aim to improve the accuracy and efficiency of state estimation~\cite{8723114,8754766}.
However, a prevalent drawback is the omission of the fundamental physical laws and constraints that govern power systems. This oversight can lead to imprecise estimates, particularly when confronted with limited data or system variations from training conditions.

Physics-informed neural networks are a class of machine learning models that integrate physical laws and constraints into the NN architecture~\cite{raissi2019physics}. By incorporating prior knowledge about the system behavior, such as conservation laws and boundary conditions, PINNs enhance the accuracy and interpretability of the learned models. This is achieved by enforcing the physics-based constraints as regularization terms during the training process, guiding the NN to produce predictions consistent with the underlying physics.

In recent years, there has been growing interest in utilizing PINNs for power system state estimation. Previous studies have explored the application of PINNs for power system parameter estimation, dynamic state estimation, and fault detection, among others~\cite{9072507,pagnier2021physics,huang2022applications}. These works have shown promising results, demonstrating the capability of PINNs to capture the complex dynamics of power systems and handle data scarcity.

This approach, using PINNs for power system state estimation offers significant advantages and paves the way for future research in this field. Firstly, it simplifies the NN architecture design by eliminating the need for system-specific designs based on system topology. This enhances flexibility and applicability across different power system configurations. Secondly, the proposed approach is not dependent on the optimal placement of PMUs or legacy sensors, making it adaptable to various measurement configurations. Lastly, it eliminates the reliance on modeling time-series dependent events, making it well-suited for real-time applications requiring accurate and timely state estimation. These advantages position the proposed approach as a valuable direction for further exploration and development in power system state estimation using PINNs.
\section{Physics-Informed Neural Networks (PINNs)}\label{sec:Methodology}
\begin{figure}[!t]
    \centering
    \includegraphics[width=0.43\textwidth]{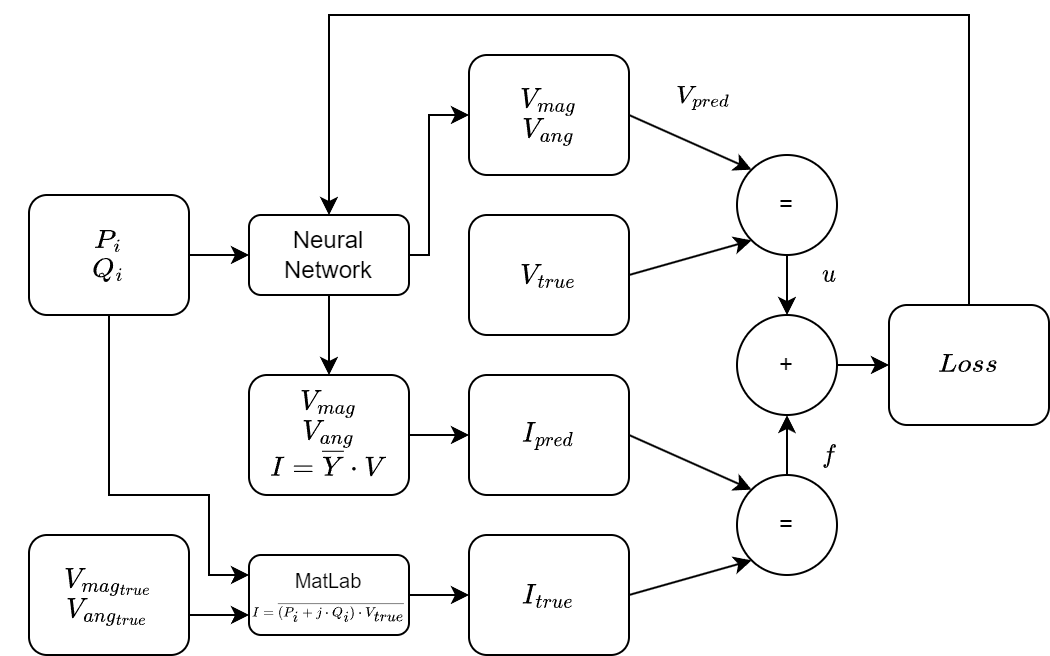}
    \vspace{-0.5\baselineskip}
    \caption{PINN training procedure dataflow diagram. A combination of \textit{data-driven} ($u$) and a \textit{physics-driven} ($f$) loss function adjusts the weights/biases of the NN in an iterative process.}
    \vspace{-\baselineskip}
    \label{fig:dataflow diagram}
\end{figure}
\subsection{Loss Function Augmentation with Physics}\label{ssec:PINNs}
The physics-derived information can be introduced to a neural network in various ways. In this paper, additional terms are introduced in the loss function that enforce the physics-based constraints during the training process. These terms can be formulated based on known relationships, equations, or laws governing the system. For example, in the context of power system state estimation, constraints related to power flow equations, Kirchhoff's laws, or the admittance matrix can be incorporated. By including these physics-based terms in the loss function, we ensure that the neural network's predictions are consistent with the physical laws and constraints of the system. During training, the network is encouraged to minimize both the discrepancy between predicted and actual values and the violation of the imposed physics-based constraints.

The loss function (hereafter $Loss$) in the proposed PINN approach is calculated as the sum of the \textit{Mean Square Error (MSE)} between the actual and inferred data ($u$) and the \textit{Mean Absolute Error (MAE)} derived from formulating a physics equation that has to be satisfied and optimally be equal to zero ($f$). Further, the terms $\lambda_1$ and $\lambda_2$ are introduced to the proposed PINN approach, which are variable weights that can be manipulated in order to change the influence of each part of the $Loss$ during the training process. The $Loss$ function can be generally expressed as,
\begin{equation} \label{eq:example loss function}
    Loss = \lambda_1 \cdot u + \lambda_2 \cdot f
\end{equation}
where $u$ represents the \textit{data-driven} part of the loss function, while $f$ is the \textit{physics-informed} part. An overview of the approach can be seen in Fig. \ref{fig:dataflow diagram}. This regularization approach helps prevent the neural network from overfitting to the training data and generating unrealistic or physically implausible results. By incorporating the physics-based constraints as regularization terms, along with variables to control their weight, a balance between data-driven learning and adherence to the underlying physics is achieved.

\subsection{Physics Formulation and Loss Function Integration}\label{ssec:formulation}
The essence of PINNs is to incorporate physical characteristics of the system into the learning process. This is achieved by adjusting the loss function and introducing a new parameter that guides the optimizer in adjusting the weights and biases of the neural network's neurons while considering the system's physics.

The task of power system state estimation involves inferring the voltage magnitude and voltage angle at each bus of the system. In this context, the regression results of the PINN are the complex voltages. The choice of using the active power ($P_i$) and reactive power injections ($Q_i$) measurements as the input dataset for the neural network is a driven by the fact that they are valuable for learning about the system's topology through data-driven patterns. The active and reactive power injections are calculated as, 
\begin{subequations}\label{eq:power flow}
\begin{equation}
    P_i = V_i \cdot \sum_{j \in N} V_j (G_{ij}\cos{\theta_{ij}} + B_{ij}\sin{\theta_{ij}})
\end{equation}
\begin{equation}
    Q_i = V_i \cdot \sum_{j \in N} V_j (G_{ij}\sin{\theta_{ij}} - B_{ij}\cos{\theta_{ij}})
\end{equation}
\end{subequations}
where, $V_i$ and $V_j$ representing the voltages magnitudes of buses i and j respectively, $\theta_{ij}$ is the difference of phase angle between buses i and j and $G_{ij}$ and $B_{ij}$ being the real and imaginary part of the admittance matrix, and $N$ represents the number of buses in a system. Based on (2a) and (2b), the net power injected in the bus is related to the complex voltage of the particular bus as well as of the buses connected to it. 

The equations also involve the admittance matrix of the power system, which contains topological and connectivity information. As shown in Eq. \eqref{eq:power flow}, the power injection is not dependent on time elements, and is instead defined as the relationship of $P_i$ and $Q_i$ as the complex loads. By learning the relationships between complex powers and voltages, the neural network can capture the underlying system topology without explicitly incorporating system-specific designs.

Similarly, the choice of utilizing current injections as the physics regularization parameter in the PINN is justified due to its inherent relationship with complex voltage, which involves the admittance matrix containing topological and connectivity information about the power system. The calculation of current injections does not introduce any time-series dependencies and can effectively serve as a physics-based constraint for the neural network. As shown in Eq. \eqref{eq:complex impedance}, which is in matrix format, there is direct relationship between the voltage ($V$) and current phasors ($I$), involving the admittance matrix ($Y$).
\begin{equation}\label{eq:complex impedance}
    \textbf{I} = \overline{\textbf{Y}} \cdot \textbf{V}
\end{equation}
By incorporating injection currents as a regularization parameter, it's ensured that the learned model adheres to the physical laws and constraints governing the power system. This choice not only promotes algorithmic stability and accuracy in the estimation process but also utilizes the topological information encoded in the admittance matrix $Y$ to further enhance the network's understanding of the power system's behavior.

The PINN is going to be trained using backpropagation, a widely used technique in training neural networks. It involves the iterative adjustment of network parameters through the minimization of a loss function. By employing backpropagation, the network propagates the error that the loss represents, backwards through the layers, hence the loop in Fig. \ref{fig:dataflow diagram}, updating the weights and biases of the neurons. This process involves calculating the gradients of the loss function with respect to the network parameters and using these gradients to adjust the parameters in a way that minimizes the $Loss$. Through repeated iterations of forward propagation, error calculation, and backpropagation, the network gradually learns to improve its predictions and minimize the $Loss$ with the aim of finding the optimal set of network parameters that minimizes the overall discrepancy between predicted and true values.

For the $u$ part of the $Loss$ function in Eq. \eqref{eq:example loss function}, the network computes its predictions in the form of $V_{mag}$ and $V_{ang}$, representing the magnitude and angle of voltage at each bus. The MSE is calculated by comparing these predictions with the ground truth values $V_{true}$. The MSE quantifies the discrepancy between the predicted and actual values, providing a measure of the network's performance, meaning that the closer the MSE is to zero, the more accurate the network's output is. This process is represented as the comparison of $V_{mag}$ and $V_{ang}$ with $V_{true}$ in Fig. \ref{fig:dataflow diagram}, resulting in the $u$ part of the $Loss$ and is derived as:
\begin{equation}
    u = Mean((V_{pred}-V_{true})^2)
\end{equation}


The $f$ part of Eq. \eqref{eq:example loss function} is derived from the comparison of the output of the neural network, $V_{mag}$ and $V_{ang}$, to the \textit{Current Injection Ground Truth} $I_{true}$ dataset as shown in Fig. \ref{fig:dataflow diagram}. This dataset is constructed from the input dataset, $Pi$ and $Qi$, and the $V_{true}$, so no new data is required to enable this procedure. $V_{mag}$ and $V_{ang}$ from the current training batch are used in conjunction with the admittance matrix $Y$, in order to generate current injection values $I_{pred}$ at each training epoch. $I_{pred}$ is then compared to the ground truth $I_{true}$, to derive their absolute difference. Therefore, the $f$ part of the $Loss$ is defined as:
\begin{equation}
    f = Mean(|I_{pred}-I_{true}|)
\end{equation}
In order to avoid scaling issues when adding two values derived from different datasets, the two parts of the loss function are re-scaled to be $\in [0,1]$:
\begin{subequations}\label{eq:normalized loss parts}
    \begin{equation}
        u_{norm} = \frac{Mean((V_{pred}-V_{true})^2)}{Max((V_{pred}-V_{true})^2)}
    \end{equation}
    \begin{equation}
        f_{norm} = \frac{Mean(|I_{pred}-I_{true}|)}{Max(|I_{pred}-I_{true}|)}
    \end{equation}
Hence, the final $Loss$ function is defined as:
    \small
    \begin{equation}\label{eq:final loss}
        Loss = \lambda_1 \cdot \frac{Mean((V_{pred}-V_{true})^2)}{Max((V_{pred}-V_{true})^2)} + \lambda_2 \cdot \frac{Mean(|I_{pred}-I_{true}|)}{Max(|I_{pred}-I_{true}|)}
    \end{equation}
    \normalsize
\end{subequations}
\section{Experimental Setup \& Results}\label{sec:Experimental Setup & Results}
\subsection{Network Hyper-parameters and Dataset Pre-processing}\label{ssec:Experimental Setup}
The experiments in this study were conducted using a neural network architecture and specific training parameters, as detailed in Table \ref{tab:hyperparameters}. Hyper-parameter tuning is beyond the scope of this work; therefore, we adopted commonly used parameters, as in literature, as a reasonable baseline for evaluation, such as a feedforward neural network using the Adam optimizer and a non-linear activation function, the hyperbolic tangent. However, future research can explore different hyper-parameter settings and conduct extensive optimization to enhance the performance and robustness of the proposed approach.

To assess the performance of the PINN, we generate two datasets representing different scenarios of the IEEE 14-bus system benchmark\cite{7151773} utilizing the PowerWorld software. The first dataset represents a steady-state condition changing the loads of the system (as in a usual system), while the second dataset captures a 20-second period encompassing the sudden shutdown of the generator at \textit{Bus 2} and the subsequent recovery period. These datasets enable the evaluation of the PINN's accuracy and algorithmic stability in estimating power system states under both normal and transient operating conditions, providing valuable insights into its performance and resilience. To ensure the accuracy of these results, a 5-fold cross-validation training strategy is employed, dividing the dataset into five subsets and performing five training iterations, where each subset served as a validation set once.

The dataset pre-processing procedure consisted of several steps to ensure optimal training of the neural network. Firstly, to avoid issues with zero division during standardization or re-scaling, any zero values in the dataset were replaced with a small non-zero value ($1e-8$). Next, in order to have realistic data, measurement noise was introduced to the simulation data to enhance the model's ability to handle variations and uncertainties in real-world scenarios. The noise levels as well as the methodology that was used for adding noise can be found in \cite{asprou2013effect}. To smooth out any anomalies in the data, the input dataset ($P_i$, $Q_i$) was standardized, e.g. normalized using the \textit{mean} and \textit{standard deviation}. Additionally, to facilitate the use of the \textit{tanh} activation function, the input dataset was re-scaled to $\in [-1, 1]$. Regarding the measurement configuration, it is implicitly assumed that real/reactive power injection measurements are available from every bus of the system. In the case of the steady state conditions datasets the measurements can be provided by the SCADA system, while in the transient operating conditions these measurements can be provided by a PMU-based system.

\begin{figure}[!t]
    \centering
    \includegraphics[width=0.5\textwidth]{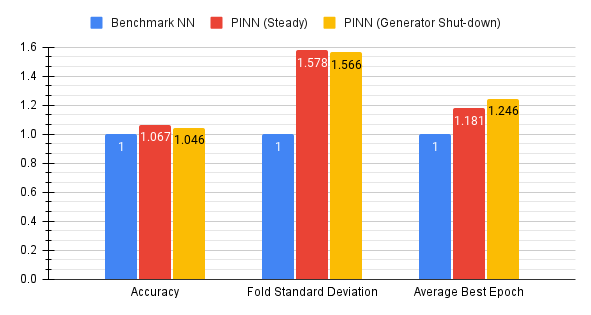}
    \vspace{-2\baselineskip}
    \caption{The PINNs' average performance, across different increment scenarios and datasets, normalized to the benchmark NN.}
    \vspace{-\baselineskip}
    \label{fig:overall performance}
\end{figure}

\begin{table}[t]
\centering
\renewcommand{\arraystretch}{1.2}
\caption{Neural Network Hyper-parameters and Setup.}
\vspace{-0.5\baselineskip}
\label{tab:hyperparameters}
\begin{tabular}{||c|c||}
\hline \hline
\textbf{Neurons/Layers} & 32 Neurons / 1 Layer \\ \hline
\textbf{Activation Function} & Hyperbolic Tangent \\ \hline
\textbf{Optimizer} & Adam \\ \hline
\multirow{4}{*}{\textbf{Dataset size}} & 192 time instances \\ & (Steady State) \\ & 2000 time instances \\ & (Generator shut-down scenario) \\ \hline
\textbf{Batch size} & 16 time instances \\ \hline
\textbf{k-Folds} & 5-fold training \\ \hline
\textbf{Epochs} & 1000 \\ \hline
\textbf{Benchmark} & 1000 \\ \hline
\textbf{Framework} & TensorFlow \\
\hline \hline
\end{tabular}
\vspace{-1.5\baselineskip}
\end{table}

\begin{table*}[t]
\centering
\renewcommand{\arraystretch}{1.2}
\caption{Steady-State Dataset Results.}
\vspace{-0.5\baselineskip}
\label{tab:steady state}
\begin{tabular}{||c||c|c||c|c||c|c||}
\hline \hline
\multirow{2}{*}{\textbf{Training Methods}} & \textbf{Cross-Validation} & \multirow{2}{*}{\textbf{Normalized}} & \textbf{Fold} & \multirow{2}{*}{\textbf{Normalized}} & \textbf{Average} & \multirow{2}{*}{\textbf{Normalized}}\\
& \textbf{Error} & & \textbf{Standard Deviation} & & \textbf{Best Epoch} & \\ \hline \hline
NN & 7.07\% & 0.00\% & 2.29\% & 0.00\% & 981 & 0.00\% \\ \hline
10\% increment & 7.55\% & 6.81\% & 1.30\% & -43.03\% & 646.2 & -34.13\% \\ \hline
20\% increment & 6.42\% & -9.19\% & 0.90\% & -60.76\% & 896.6 & -8.60\% \\ \hline
25\% increment & 6.43\% & -9.02\% & 0.94\% & -58.83\% & 875.4 & -10.76\% \\ \hline
33\% increment & 6.32\% & -10.62\% & 1.12\% & -50.82\% & 866.8 &-11.64\% \\ \hline
50\% increment & 6.27\% & -11.26\% & 0.89\% & -60.89\% & 733 & -25.28\% \\
\hline \hline
\end{tabular}
\end{table*}

\begin{table*}[t]
\centering
\renewcommand{\arraystretch}{1.2}
\caption{Generator Shut-down Dataset Results.}
\vspace{-0.5\baselineskip}
\label{tab:generator shutdown}
\begin{tabular}{||c||c|c||c|c||c|c||}
\hline \hline
\multirow{2}{*}{\textbf{Training Methods}} & \textbf{Cross-Validation} & \multirow{2}{*}{\textbf{Normalized}} & \textbf{Fold} & \multirow{2}{*}{\textbf{Normalized}} & \textbf{Average} & \multirow{2}{*}{\textbf{Normalized}}\\
& \textbf{Error} & & \textbf{Standard Deviation} & & \textbf{Best Epoch} & \\ \hline \hline
NN & 5.41\% & 0.00\% & 0.64\% & 0.00\% & 877.6 & 0.00\% \\ \hline
10\% increment & 5.38\% & -0.58\% & 0.54\% & -15.29\% & 695.8 & -20.72\% \\ \hline
20\% increment & 5.19\% & -4.19\% & 0.18\% & -72.41\% & 683.8 & -22.08\% \\ \hline
25\% increment & 5.09\% & -5.98\% & 0.30\% & -53.27\% & 603.6 & -31.22\% \\ \hline
33\% increment & 5.11\% & -5.69\% & 0.16\% & -75.63\% & 732.4 & -16.55\% \\ \hline
50\% increment & 5.07\% & -6.32\% & 0.22\% & -66.28\% & 593 & -32.43\% \\
\hline \hline
\end{tabular}
\vspace{-1.5\baselineskip}
\end{table*}

\subsection{Results \& Discussion}\label{ssec:results}
The results for the two scenarios, namely the ``steady state'' and ``generator shut-down'', are presented in Table \ref{tab:steady state} and Table \ref{tab:generator shutdown}, respectively. The results are presented in both individual and in normalized format, in order to show the relative performance of each experiment to the baseline NN as a percentage change. Different training regimes are employed by adjusting the values of $\lambda_1$ and $\lambda_2$, as defined in Section \ref{ssec:formulation}, Eq. \eqref{eq:final loss}, to balance the data-driven and physics-informed aspects of the learning process. The training regimes, represented by the ``$\%$ \textit{increment}'' scenarios, involved gradually decreasing $\lambda_1$ and increasing $\lambda_2$ at specific intervals during the training epochs. For example, in the ``$10\%$ \textit{increment}'' scenario, $\lambda_1$ decreased by $10\%$ and $\lambda_2$ increased by $10\%$ every $100$ epochs. This progressive adjustment allows to prioritize data-driven learning in the initial stages of training and gradually focus on leveraging the physics-informed constraints to fine-tune the model's performance. By carefully controlling these parameters, the learning process was optimized and the PINN has improved accuracy over the plain NN. To show the improved accuracy of the PINN over the NN, the results in Table \ref{tab:steady state} and Table \ref{tab:generator shutdown} are averaged over all the training methods and plotted in the first three bars for the two datasets as seen in Fig. \ref{fig:overall performance}. The accuracy for the state steady scenario is improved by $6.7\%$ on average, and by $4.6\%$ on the generator shut-down dataset.

By employing a 5-fold cross-validation training strategy enabled us to assess the similarity and sensitivity of each approach to variations in the datasets. To evaluate the algorithmic stability of the training process, we calculate the standard deviation of validation error across the population of folds, providing a quantitative measure of the consistency of the results obtained from different training folds. This analysis helped us evaluate the generalization capability of the PINN model and provided insights into the model's sensitivity to changes in the training data, showing that introducing physics-based regularization in the training process, greatly benefited the resulting neural network across all different scenarios of weight adjustments. The mean standard deviation between the training folds for the steady state dataset has been improved by $57.8\%$ and by $56.6\%$ for the generator shut-down dataset.

At each fold training process, the epoch that yielded the best validation error is noted. This allows to identify the epoch at which the model achieved its maximum performance, pinpointing the optimal point within the 1000 training epochs. The ability to reach peak performance earlier in the training process indicates that the physics-enhanced model requires less training effort and can expedite the convergence to the desired accuracy. The PINN has reached peak performance $18.1\%$ and $24.6\%$ faster than the NN within the 1000 epoch training regime of the two datasets, as shown in Fig. \ref{fig:overall performance}.
\section{Conclusion \& Future Work}\label{sec:Conclusion}
In conclusion, we proposed a novel PINN approach for power system state estimation. Extensive experiments demonstrated that the proposed PINN achieved higher accuracy, required less training effort, and demonstrated improved algorithmic stability in regression results compared to a benchmark NN in the state estimation of power systems under high availability of measurements.

In future work, we will investigate the application of PINNs in limited observability scenarios and conduct hyperparameter tuning to optimize their performance. Additionally, we will explore the possibility of PINNs being better than NNs when a dataset that includes faults is used, as they can incorporate physical constraints into the training process.

\section*{Acknowledgment}\label{s:Acknowledgement}
This work was partially supported by the European Union’s Horizon 2020 research and innovation programme under grant agreement No 101016912 (ELECTRON), and by the European Union’s Horizon 2020 research and innovation programme under grant agreement N0 739551 (KIOS CoE – TEAMING) and from the Republic of Cyprus through the Deputy Ministry of Research, Innovation and Digital Policy. This publication is partially based upon work supported by King Abdullah University of Science and Technology (KAUST) under Award No. ORFS-CRG11-2022-5021.
\vspace{-0.5\baselineskip}

\bibliographystyle{IEEEtran}
\bibliography{sources}

\end{document}